\documentclass{article} 
\usepackage{iclr_r2fm,times}
\usepackage{multirow}
\usepackage{booktabs}    
\usepackage{bbding}
\usepackage{siunitx}
\usepackage{graphicx}
\usepackage{color}
\usepackage{longtable}
\usepackage{caption}
\usepackage{soul}
\usepackage[para]{footmisc}
\usepackage{xcolor}
\usepackage[most]{tcolorbox}
\usepackage{colortbl}
\usepackage{lipsum}

\usepackage{ulem}
\usepackage{amsmath,amsfonts,bm}

\def\eqref#1{equation~\ref{#1}}

\def\1{\bm{1}}

\DeclareMathAlphabet{\mathsfit}{\encodingdefault}{\sfdefault}{m}{sl}
\SetMathAlphabet{\mathsfit}{bold}{\encodingdefault}{\sfdefault}{bx}{n}

\newcommand{\eg}{\textit{e.g., }}

\newcommand\redsout{\bgroup\markoverwith{\textcolor{red}{\rule[0.4ex]{2pt}{1pt}}}\ULon}

\usepackage{hyperref}

\definecolor{gray}{HTML}{f1f3f5}
\tcbset{sharp corners, colback = white, before skip = 0.2cm,  after skip = 0.5cm   }   
\newtcolorbox{custombox}{left=1mm,right=1mm, top=0.5mm, bottom=0.5mm, colback = gray, boxrule = 0.1pt  }

\usepackage{url}

\title{\ours{} : Revising after Explanation reduces the Factual Errors in LLM Responses
}

\author{Juyeon Kim$^1$ \hspace{4mm} Jeongeun Lee$^1$\thanks{~Equal contribution.}\hspace{4mm}Yoonho Chang$^1$\footnotemark[1] \\ \textbf{Chanyeol Choi}$^2$\hspace{4mm}  \textbf{Junseong Kim}$^2$\thanks{~Corresponding authors.}\hspace{4mm}  \textbf{Jy-yonh Sohn}$^1$\footnotemark[2]\vspace{-3mm}
\\\\\text{$^1$Yonsei University \hspace{4mm} $^2$Linq} \vspace{1mm}
\\ \texttt{\{culy1125, ljeadec31, yoonho990625, jysohn1108\}@yonsei.ac.kr}
\\ \texttt{\{jacob.choi, junseong.kim\}@getlinq.com}}

\newcommand{\ours}{\textsc{Re-Ex}}
\iclrfinalcopy 
\begin{document}
\maketitle

\begin{abstract}

Mitigating hallucination issues is a key challenge that must be overcome to reliably deploy large language models (LLMs) in real-world scenarios. Recently, various methods have been proposed to detect and revise factual errors in LLM-generated texts, in order to reduce hallucination.
 In this paper, we propose \ours{}, a method for post-editing LLM-generated responses. \ours{}  introduces a novel reasoning step dubbed as the factual error \textit{explanation step}.  \ours{} revises the initial response of LLMs using 3-steps : first, external tools are used to retrieve the evidences of the factual errors in the initial LLM response; next, LLM is instructed to \textit{explain} the problematic parts of the response based on the gathered evidence; finally, LLM revises the initial response using the \textit{explanations} provided in the previous step. In addition to the \textit{explanation step}, \ours{} also incorporates new prompting techniques to reduce the token count and inference time required for the response revision process. Compared with existing methods including FacTool, CoVE, and RARR, \ours{} provides better detection and revision performance with less inference time and fewer tokens in multiple benchmarks. The code for \ours{} is available at: \url{https://github.com/juyeonnn/ReEx}.

\end{abstract}

\section{Introduction}
\vspace{-1mm}

Large Language Models (LLMs) exhibit remarkable text generation abilities and deep language understanding. However, a significant challenge persists: the reliability of LLM-generated text is often compromised due to frequent instances of hallucinations~\citep{ji2023survey}. Even widely used language models, including ChatGPT, suffer from this issue. This unreliability severely hinders adoption in real-world scenarios where accurate, up-to-date information is essential. Therefore, detecting and correcting factual errors in LLM-generated text is crucial.
To tackle this issue, there have been some approaches which instructed LLMs to self-refine their responses~\citep{tonmoy2024comprehensive}. For example, the authors of \citep{manakul2023selfcheckgpt,mündler2023selfcontradictory} focus on detecting \textit{self-contradiction} hallucination, while other methods extended the reasoning steps to reduce hallucination\citep{dhuliawala2023chainofverification,zhao2023verifyandedit}. However, as discussed in~\citep{mallen2023trust}, relying solely on their internal knowledge base still has a probability of generating texts that contain factual errors.

While Retrieval-Augmented Generation (RAG) ~\citep{NEURIPS2020_6b493230} based methods mitigate the aforementioned challenges by leveraging information retrieved from external sources, a significant challenge persists: effectively retrieving and processing the necessary information as evidence. Determining how to handle or segment content, given the length and complexity of both the original response and gathered evidence, becomes crucial.
\citet{gao2023rarr} proposes a method that iterates through the evidence one by one to check whether the evidence aligns with the response, and revises the response accordingly. Several fact-checking methods such as~\citep{chern2023factool} and~\citep{li2023selfchecker} focus on factual error detection by first splitting the response into claims and then generating queries to verify each claim. However, both iterative evidence checking and claim-by-claim verification can be time-consuming and costly, limiting their use in real-world scenarios.

In this paper, we propose \ours{}, which aims to post-edit factual errors in LLM-generated responses using evidence gathered from external sources. Building on research demonstrating significant LLM reasoning improvement through \textit{intermediate reasoning steps}~\citep{wei2023chainofthought,deng2023rephrase,chowdhery2022palm}, \eg chain-of-thought prompting, we introduce an \textit{explanation step} to reduce hallucination errors. \ours{} first gathers evidences of factual errors in the response, and then \textit{explains} the factual errors in the response, and finally revise the response accordingly (as shown in Figure~\ref{fig:example}). Our contributions are significant in three primary aspects:
\begin{enumerate}
\item Instead of making LLMs \textit{directly} revise the response, \ours{} employs the process of \textit{revising after explanation}: we let LLMs first \textit{explain} the factual errors (based on the evidences), and then let LLMs revise their response accordingly. This intermediate \textit{explanation step} leads to better revision performance, as confirmed by our ablation study.

\item  Compared with baselines including FacTool~\citep{chern2023factool}, CoVe~\citep{dhuliawala2023chainofverification} and RARR~\citep{gao2023rarr}, our method \ours{} enjoys  higher revision performances in multiple benchmarks (FactPrompt~\citep{chern2023factool}, WiCE~\citep{kamoi2023wice}) and in the annotated dataset given in~\citep{min2023factscore}.

\item \ours{} enhances efficiency, achieving up to 10x faster processing and 6x fewer tokens, by integrating certain steps with better prompting. Instead of generating search queries after splitting the response into claims, \ours{} employs a single-step process directly prompting LLM to decompose text and generate sub-questions simultaneously (inspired by the success of question decomposition method~\citep{radhakrishnan2023question}). Additionally, rather than iterative evidence checking, \ours{} presents all evidence at once, instructing LLMs to \textit{explain} factual errors.
\end{enumerate}

\begin{figure}[t]{
\vspace{-8mm}
\begin{custombox}
\small
\textbf{Prompt $ P $}: Which country or city has the maximum  number of nuclear power plants? \\
\textbf{Initial Response $ R_{\text{initial}} $}  : \underline{The United States} has the highest number of nuclear power plants in the world, with \underline{94 operating reactors}. Other countries with a significant number of nuclear power plants ...
\end{custombox}
\vspace{-5.5mm}

\begin{custombox}
\small
\textbf{Sub-Question $Q_{1}$} : \underline{How many operating reactors} does the United States have?  \\
\textbf{Sub-Answer $A_{1}$}   : ... the United States had \underline{93 operating commercial reactors} at 54 nuclear power plants ...
\end{custombox}
\vspace{-5.5mm}

\begin{custombox}
\small
\textbf{Explanation of Factual Error $E_1$} : 
     {The initial response states that the United states has \textcolor{red}{\textbf{94 operating reactors}}, but the correct number is \textcolor{blue}{\textbf{93 operating reactors} }.  }
\end{custombox}
\vspace{-5.5mm}

\begin{custombox}
\small
\textbf{Revised Response $ R_{\text{revised}}$} : 
     The United States has the highest number of nuclear power plants in the world, with \redsout{\textcolor{red}{94 operating reactors}} \textcolor{blue}{\textbf{93 operating reactors} }. Other countries with a significant number of ...
\end{custombox}
\vspace{-1mm}
}
\vspace{-2mm}
\caption{An example of revising the response with \ours{}. The initial response $R_{\text{initial}}$ states the US has 94 operating reactors, proven inaccurate by search result $A_1$. \ours{} \textit{explains} the factual error in $E_1$ and revises the response accordingly.} 
\vspace{-1mm}
\label{fig:example}
\vspace{-4mm}
\end{figure}

\vspace{-3mm}
\section{Method}
\vspace{-1mm}

Suppose a LLM generates its initial response $ R_{\text{initial}} $ given a prompt $ P $. \textit{\ours{}} revises this response in three steps. In step 1, the LLM generates $ N $ sub-questions $ \{Q_{i}\}_{i=1}^N $ related to the prompt and the initial response. External search tools are then used to obtain answers for each question, denoted by $ \{A_{i}\}_{i=1}^N $, thus forming $ N $ evidence pairs $ \{(Q_{i}, A_{i})\}_{i=1}^{N} $. In step 2, the LLM is instructed to \textit{explain} each factual error in the response using the evidence pairs $ \{(Q_{i}, A_{i})\}_{i=1}^{N} $. The $ M $ \textit{explanations} of factual errors produced in this stage are denoted by $ \{E_j\}_{j=1}^{M} $. Finally, in step 3, \textit{\ours{}} produces the final revised response $ R_{\text{revised}} $ using $ \{E_j\}_{j=1}^{M} $, yielding a more factually accurate and refined response with corrected factual errors. The overview of \ours{} is in Fig.\ref{fig:ReEx}, and the prompt used in our method is in Appendix \ref{app:prompt}. Below we provide a detailed explanation on each step:

\paragraph{Step 1. Evidence Retrieval}
\ours{} first decomposes the initial response $ R_{\text{initial}} $, along with the given prompt $ P $, into a series of self-contained, concise sub-questions. The prompt used for generating sub-question is given in Table~\ref{tab:query-generation} in Appendix~\ref{app:query}.  Using the sub-question $Q_{i}$ as search queries, relevant snippets  such as as answer boxes, Knowledge Graph, and organic search results are retrieved using Google Search\footnote{\url{https://serper.dev}}. These search results are denoted as sub-answer $A_{i}$ for each sub-question $Q_{i}$.

\vspace{-1mm}
\paragraph{Step 2. Factual Error Explanation}
Using the pairs of sub-question and sub-answer, denoted as $ \{(Q_{i}, A_{i})\}_{i=1}^{N} $, \ours{} employs \textit{explanation step} to pinpoint and explain each factual error. The \textit{explanation} $E_i$ of each factual error includes both the factual error in the initial response $ R_{\text{initial}} $ and the factually accurate information, which corrects the error, retrieved from $ \{(Q_{i}, A_{i})\}_{i=1}^{N} $.
The prompt used for factual error explanation is given in Table~\ref{tab:explanation}.

\vspace{-1mm}
\paragraph{Step 3. Revision}
The revised response $ R_{\text{revised}} $ is generated using the explanation $ \{E_i\}_{i=1}^{M} $ of the factual error to revise the initial response $ R_{\text{initial}} $. Two prompting options are evaluated: \textit{one-step}, where both the revision step and the explanation step are conducted within a single prompt, and \textit{two-step}, where both steps are conducted with two separate prompts. The prompt used for the first option is in Table~\ref{tab:one-step} and the prompts for the second option is in Tables~\ref{tab:explanation} and ~\ref{tab:revision}.

\begin{figure*}[t]
    \begin{center}
    \vspace{-12mm}
    \includegraphics[scale=0.5]{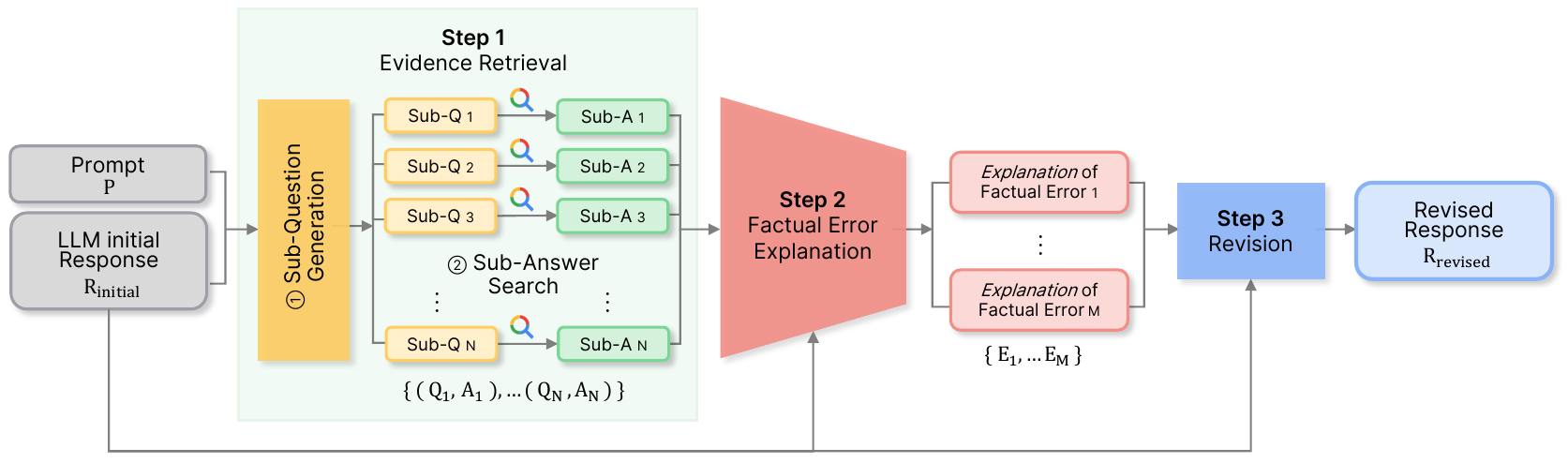}  
    \end{center}  
    \vspace{-3mm}
    \caption{Overview of \ours{}, which revises the initial response $R_{\text{initial}}$ of LLMs. First, \ours{} retrieves the evidences for factual errors in $R_{\text{initial}}$ by using external tools; it first generates the sub-questions $\{Q_i\}_{i=1}^N$ useful for checking the factual errors and then gets the answers (or evidences) $\{A_i\}_{i=1}^N$ from external sources. Second, \ours{} lets LLMs explain the factual errors in $R_{\text{initial}}$ based on the evidences, thus getting the explanations $\{E_i\}_{i=1}^M$. Finally, LLM revises its response based on the explanations, outputting the revised response $R_{\text{revised}}$.}
    \label{fig:ReEx}
    \vspace{-3mm}
\end{figure*}

\vspace{-2mm}
\section{Experiments}
\vspace{-1mm}

\subsection{Experimental Setup}\label{sec:exp-setup}

\paragraph{Baselines.} 
We compare \ours{} with three baselines: CoVE~\citep{dhuliawala2023chainofverification} refines responses through self-verification, tested in factored (CoVE \textit{(F)}) and factor+revise (CoVE \textit{(F+R)}) variants; RARR~\citep{gao2023rarr} which \textit{revises} responses via an external search method; and FacTool~\citep{chern2023factool} which employs external searches for detailed factuality \textit{detection}. Notably, FacTool is limited to factual error detection, while CoVE \textit{(F)} focuses solely on revision without providing response-level consistency labels. The comparison of baseline methods is provided in Table \ref{tab:baseline-comparison} and more details are available in Appendix \ref{app:baseline}.
\vspace{-1mm}
\paragraph{Detection Task.}
For \ours{} and baselines, we evaluate the ability to identify (or detect) factual errors at the response level using two distinct benchmarks. First, we utilize the FactPrompt dataset~\citep{chern2023factool}, which includes 50 real-world prompts and corresponding responses from GPT-3.5. Additionally, we employ the WiCE dataset \citep{kamoi2023wice}, containing 358 claims with multiple pieces of information. The balanced accuracy and F1 score are chosen as evaluation metrics. To accommodate varying labeling criteria across baselines and datasets, we transformed the data into binary labels. Further details are available in Appendix \ref{app:dataset}.

\vspace{-1mm}
\paragraph{Revision Task.}
For \ours{} and baselines, we evaluate the ability to revise the response if it has factual errors. 
We test on 157 responses from GPT-3.5, each containing an average of 31.1 labeled fact units. These fact units consist of concise sentences, each conveying a single piece of information, and are annotated by humans in a paper proposing FActScore~\citep{min2023factscore}. 
Let a response $R_{\text{initial}}$ has $n$ fact units, while $n_f$ of them are false.
Let $n_{ft}$ be the number of fact units that are incorrectly stated in $R_{\text{initial}}$, while correctly written in $R_{\text{revised}}$. Similarly, let $n_{tt}$ be the number of fact units that are correctly written in both $R_{\text{initial}}$ and $R_{\text{revised}}$.
We use two metrics, (1) \textit{correction accuracy} = $\frac{n_{ft}}{n_f}$ and (2) \textit{revision accuracy} = $\frac{n_{ft} + n_{tt}}{n}$; the correction accuracy measures the ratio of \textit{corrected} fact units that are incorrectly stated in the initial response, and the revision accuracy measures the ratio of the correct fact units in the \textit{revised} response. Here, $n_{ft}$ and $n_{tt}$ are measured by using NLI model introduced in~\citep{liu2023evaluating}, to determine whether they \textit{entail}, are \textit{neutral}, or \textit{contradict} the revised response. Further details are provided in Appendix ~\ref{sec:rev-corr-score-details}.

\subsection{Experimental Results}

\begin{table}[t]

\vspace{-10mm}
    \centering
    \small
    \setlength\tabcolsep{5pt}
    \begin{tabular}{@{}rrccccccccc@{}}
        \toprule
        
        & & \multicolumn{2}{c}{\textbf{BAcc. } \scriptsize{(↑)}} 
        & \multicolumn{2}{c}{\textbf{F1 } \scriptsize{(↑)}} 
        & \multicolumn{2}{c}{\textbf{Avg.Time} \scriptsize{(sec, ↓)}} 
        & \multicolumn{2}{c}{\textbf{Avg.Token} \scriptsize{(K, ↓)}} \\
        \cmidrule(r){3-4}
        \cmidrule(r){5-6}
        \cmidrule(r){7-8}
        \cmidrule(r){9-10}
        Dataset &Methods & \multicolumn{1}{c}{\scriptsize{GPT-3.5}} & \multicolumn{1}{c}{\scriptsize{GPT-4}} &\multicolumn{1}{c}{\scriptsize{GPT-3.5}} & \multicolumn{1}{c}{\scriptsize{GPT-4}}  & \multicolumn{1}{c}{\scriptsize{GPT-3.5}} & \multicolumn{1}{c}{\scriptsize{GPT-4}}  & \multicolumn{1}{c}{\scriptsize{GPT-3.5}} & \multicolumn{1}{c}{\scriptsize{GPT-4}}  \\
        \midrule

         \multirow{4}{*}{\small{Fact}} & FacTool & 66.5 & 68.0  & 52.6  & 57.1  & 20.0 & 26.2 & 3.1K & 4.1K\\ 
         \multirow{4}{*}{\small{Prompt}}& CoVE \small{\textit{(F+R)}} & 62.0 & 54.2 & 59.6 & 53.1 & 5.7 & 23.3 & 6.0K & 5.9K \\ 
          & RARR & 63.0 & 71.0 & 57.0 & 68.0 & 59.9 & 119.7 & 3.2K & 3.1K   \\
            \cmidrule(r){2-10}
        & \textbf{\ours{}} \small{\textit{(one-step)}}  & \textbf{75.9} & \underline{84.2} & \underline{73.9} & \underline{83.3} & \underline{4.8} & \textbf{19.0} & \textbf{1.0K}  & \textbf{1.0K} \\ 
         & \textbf{\ours{}} \small{\textit{(two-step)}} & \underline{74.7} & \textbf{87.0} & \textbf{75.5} & \textbf{86.8} &  \textbf{4.8}  &\underline{19.1} &  \underline{1.0K} & \underline{1.1K} \\  

        \midrule  
         \multirow{5}{*}{\small{WiCE}}& FacTool & 50.8& \underline{55.3} & 22.2 & 35.8 & 6.5 & 25.3 & 2.7K & 2.9K  \\ 
         & CoVE \small{\textit{(F+R)}}  &51.8  & 49.3 & 22.5 & 3.3 & 7.3 & 22.4 & 5.2K & 5.0K \\ 
         & RARR & \textbf{56.0} & 53.0 & \textbf{46.0} & \underline{43.0} &  59.1 & 112.6 & 2.8K & 2.8K\\
        
         \cmidrule(r){2-10}
          
         & \textbf{\ours{}} \small{\textit{(one-step)}} & 48.5 & 50.7 & 33.1 & 41.8 &\textbf{5.4} & \underline{17.8} & \textbf{0.8K} & \textbf{0.8K}  \\
         & \textbf{\ours{}} \small{\textit{(two-step)}}& \underline{52.8} & \textbf{55.8}  &  \underline{39.2} & \textbf{47.6}  &  \underline{5.6} & \textbf{15.4} & \underline{0.9K} & \underline{0.8K} \\ 
         \bottomrule
    
    \end{tabular}
    \vspace{-1mm}
    \caption{Experimental results on the task of detecting factual errors. Compared with other baselines, \ours{} has higher balanced accuracy (BAcc) and F1 score, by using less time and tokens.
    } 
    
    \label{tab:experiment2}
\end{table}

\begin{table*}[t]
    \small
    \centering

    \begin{tabular}{@{}rcccccccccccc@{}}
        \toprule 

        & \multicolumn{2}{c}{\textbf{Correction Acc.} \scriptsize{(↑)}}   
        & \multicolumn{2}{c}{\textbf{Revision Acc.} \scriptsize{(↑)}}   
        & \multicolumn{2}{c}{\textbf{Avg.Time} \scriptsize{(sec, ↓)}} 
        & \multicolumn{2}{c}{\textbf{Avg.Token} \scriptsize{(K, ↓)}} \\
        
        \cmidrule(r){2-3}
        \cmidrule(r){4-5}
        \cmidrule(r){6-7}
        \cmidrule(r){8-9}

         Methods & \multicolumn{1}{c}{\scriptsize{GPT-3.5}} & \multicolumn{1}{c}{\scriptsize{GPT-4}} 
         &\multicolumn{1}{c}{\scriptsize{GPT-3.5}} & \multicolumn{1}{c}{\scriptsize{GPT-4}}  
         &\multicolumn{1}{c}{\scriptsize{GPT-3.5}} & \multicolumn{1}{c}{\scriptsize{GPT-4}} 
         & \multicolumn{1}{c}{\scriptsize{GPT-3.5}} & \multicolumn{1}{c}{\scriptsize{GPT-4}} \\ 
         \midrule

        RARR &	36.9 & 39.1  
        & 63.8 & 68.7
        & 106.9 & 257.9  & 3.7& 3.9   \\
        CoVE \small{\textit{(F)}} &  36.3  & 33.6 
        & 64.9 &  69.9
         & \textbf{8.7}& \textbf{36.1} & 3.0 & 3.0  \\
        CoVE \small{\textit{(F+R)}}& 35.9 & 19.4 
        & 62.0 & 67.0
      & 16.2 & 49.8 & 13.0 & 13.1 \\ 
        \midrule
      \textbf{\ours{}} \small{\textit{(one-step)}}  &  \underline{42.8}  &  \underline{46.6} 
        & \underline{70.3} & \textbf{75.4}
        & \underline{11.4} &\underline{48.6} & \textbf{2.1} &  \textbf{2.4} \\
         \textbf{\ours{}} \small{\textit{(two-step)}}& \textbf{52.2} & \textbf{50.2} 
        & \textbf{71.5} & \underline{74.9} 
         & 12.5 & 48.9 & \underline{2.7}  &  \underline{2.8}  \\ 
    \bottomrule
    \end{tabular}
    \vspace{-1mm}
    \caption{Experimental result on the task of revising factual errors, tested on the dataset given in~\citep{min2023factscore}.
    \ours{} enjoys the best  accuracies with reduced time/tokens, compared with baselines. 
    }
    \label{tab:experiment3}
     \vspace{-3mm}
\end{table*}

\begin{table}[t]
\vspace{-10mm}
    \centering
    \small
    \setlength\tabcolsep{5pt}
    \begin{tabular}{@{}ccccccc@{}}

        \toprule
         \multicolumn{1}{c}{\small{Step 1}}
         & \multicolumn{1}{c}{\small{Step 2}}
         & \multicolumn{1}{c}{\small{Step 3}}
        & \multicolumn{2}{c}{\textbf{Correction Acc.} \small{(↑)}}   
        & \multicolumn{2}{c}{\textbf{Revision Acc.} \small{(↑)}} \\
        \cmidrule(r){1-1}
        \cmidrule(r){2-2}
        \cmidrule(r){3-3}
        \cmidrule(r){4-5}
        \cmidrule(r){6-7}

        \small{Evidence} & \small{Explanation} & \small{Revision}
         &\multicolumn{1}{c}{\scriptsize{GPT-3.5}} & \multicolumn{1}{c}{\scriptsize{GPT-4}}  
         &\multicolumn{1}{c}{\scriptsize{GPT-3.5}} & \multicolumn{1}{c}{\scriptsize{GPT-4}}  \\
        \midrule 
         - & - & $\checkmark$ & 8.4 & 24.5 & 62.2 & 67.7\\
        \midrule
         - & $\checkmark$ & $\checkmark$  & 28.7 \scriptsize{+20.3\%p} & 26.7 \scriptsize{+2.2\%p} & 65.4 \scriptsize{+2.2\%p}& 70.9 \scriptsize{+3.2\%p}\\
         
         $\checkmark$ $\cdot$ \small{Internal} & - & $\checkmark$  & \underline{40.3} \scriptsize{+31.9\%p}  & 31.8 \scriptsize{+7.3\%p} &  67.7 \scriptsize{+5.5\%p} & 66.7 \scriptsize{-1.0\%p}\\

         $\checkmark$ $\cdot$ \small{Internal} & $\checkmark$  & $\checkmark$  & 39.9 \scriptsize{+31.5\%p} & 35.5 \scriptsize{+11.0\%p} & \underline{68.8} \scriptsize{+6.6\%p} & 67.5 \scriptsize{-0.2\%p}\\

         $\checkmark$ $\cdot$ \small{External}& - & $\checkmark$  & 25.4 \scriptsize{+17.0\%p}& \underline{45.2} \scriptsize{+20.7\%p}& 67.5 \scriptsize{+7.0\%p}& \underline{74.7} \scriptsize{+7.0\%p} \\

        $\checkmark$ $\cdot$ \small{External}& $\checkmark$ & $\checkmark$  & \textbf{52.2} \scriptsize{+43.8\%p}& \textbf{50.2} \scriptsize{+25.7\%p}& \textbf{71.5} \scriptsize{+9.3\%p}& \textbf{74.9} \scriptsize{+7.2\%p}\\ 

        \bottomrule
    \end{tabular}
    \vspace{-1mm}
    \caption{Ablation Study of each stage of \ours{}, tested on the FactScore dataset used in Table \ref{tab:experiment3}. We tested the revision task on GPT-3.5 and GPT-4, where steps 1, 2, and 3 are illustrated in Fig.~\ref{fig:ReEx}. Excluding step 1 or step 2 results in over a 20\% degradation in correction accuracy for GPT-3.5. We also tested the effect of using external sources in step 1, by comparing with a variant of our method using internal source (i.e., retrieve from the response of LLMs to the query), similar to CoVE~\citep{dhuliawala2023chainofverification} shown in Table \ref{tab:baseline-comparison} in Appendix.}
    \label{tab:ablation1}
    \vspace{-5mm}
\end{table}

\vspace{-1mm}
\paragraph{\ours{} performs well with low cost.
}
As in Table~\ref{tab:experiment2}, \ours{} achieves the best scores (balanced accuracy and F1 score) on the detection of factual errors.  In particular, the performance improvement for GPT-4 was more pronounced, with an F1 score of 86.8 compared to 53.1 for CoVE \textit{(F+R)} in the FactPrompt dataset. This is a much larger margin than that observed in GPT-3.5, where \ours{} achieved an F1 of 75.5 against 52.6 for FacTool. As in Table~\ref{tab:experiment3},
\ours{} far outperforms existing baselines in the revision task, for both GPT-3.5 and GPT-4 models. 
For example, the correction accuracy increases by more than 15\% for GPT-3.5, and by more than 10\% for GPT-4. Similarly, the revision accuracy increases by more than 5\% for both GPT-3.5 and GPT-4. As in Tables~\ref{tab:experiment2} and~\ref{tab:experiment3}, \ours{} requires much less time and tokens than baselines to achieve such high performance, \eg \ours{} uses 6x fewer tokens than CoVE \textit{(F+R)}. Table~\ref{tab:revision-example1} and ~\ref{tab:revision-example2} in Appendix compares the revised responses for different methods including \ours{}, CoVE and RARR. One can confirm that \ours{} successfully corrects the factual errors in the initial response, while CoVE and RARR cannot.

\vspace{-1mm}
\paragraph{When and why does \ours{} perform well?}
Recall that our method has two versions: \ours{} \textit{(two-step)} which uses separate prompts for explanation and revision, and \ours{} \textit{(one-step)} which uses a single prompt for both. As shown in Tables~\ref{tab:experiment2} and~\ref{tab:experiment3}, \ours{} \textit{(two-step)} outperforms \ours{} \textit{(one-step)} in most cases. The correction accuracy gap is around 10\%p, suggesting that splitting the task into smaller steps is helpful, similar to the observation reported in~\cite{bubeck2023sparks}. Furthermore, \ours{} demonstrates more consistent and reliable performance compared to baseline methods. As shown in Table ~\ref{tab:baseline-comparison} in Appendix, baselines can exhibit significant variation and vulnerability based on few-shot examples or labeled data, as discussed in ~\cite{nguyen2023incontext,lu2022fantastically}. In contrast, \ours{} enables a zero-shot setting with improved prompting techniques.

\vspace{-1mm}
\paragraph{Ablation Study.} Recall that \ours{} consist of 3 steps as shown in Fig.~\ref{fig:ReEx}.  As in Table~\ref{tab:ablation1}, ablating each step significantly reduces the correction accuracy for GPT-3.5. This shows that both the evidence retrieval step and the factual error explanation step are necessary for successful revision. 
In addition, the results in Table~\ref{tab:sub-answer-example} shows that utilizing external evidence outperformed relying solely on internal response. This result coincides with our qualitative comparison in Table~\ref{tab:sub-answer-example} in Appendix, showing that  relying solely on the internal knowledge of LLMs can lead to insufficient evidence since they often provide outdated information or the document that does not contain the data we want to retrieve.
\vspace{10mm}

\vspace{-10mm}
\section{Related Works}
\vspace{-1mm}
\paragraph{Fact Verification.}
The task of fact verification, which aims to assess the factual accuracy of claims, has seen significant research interest due to its wide applications. Existing methodologies for fact verification have often focused on short-form datasets, primarily FEVER~\citep{thorne2018fever} or VitaminC~\citep{schuster2021vitamin}, which often struggle with the length and complexity of LLM-generated texts. Consequently, researchers have begun exploring ways to detect factual errors in longer textual outputs~\citep{fan2020generating, lattimer2023fast, soleimani2023nonfacts}.
The research also explored post-editing methods that could provide revised sentences along with the output of factual error detection, in order to achieve more practical aims. Drawing inspiration from the recent success of LLMs by using the Chain-of-Thought (CoT) method \citep{wei2023chainofthought}, some methods have tried to revise factual error via an additional reasoning step \citep{dhuliawala2023chainofverification,zhao2023verifyandedit}. Following this line of research, our work introduces an explanation step as an intermediate reasoning process, which does not depend on label or few-shot examples (see Table \ref{tab:baseline-comparison} in Appendix).

\vspace{-1mm}
\paragraph{Tool Use.}
Language models inherently possess a limited knowledge base derived from their pre-trained data. This limitation has motivated the recent incorporation of various tools to augment these models, significantly enhancing their capabilities~\citep{schick2023toolformer, zhuang2023toolqa, shen2023hugginggpt, liang2023taskmatrix}.
While some methods utilize established knowledge repositories like Wikipedia~\citep{semnani2023wikichat,li2023selfchecker,yu2023chain}, others leverage search APIs to access more current and broader information~\citep{gou2023critic,chern2023factool,gao2023rarr}. Our method also employs the Google Search API to address the limitations of pre-existing knowledge databases and enhance its suitability for more practical scenarios.

\vspace{-1mm}
\paragraph{Automated Evaluation.}
Various metrics like BLEU~\citep{papineni2002bleu} and ROUGE~\citep{lin2004rouge} have been used to assess machine-generated text quality. Recent methods have leveraged LLMs for automated text evaluation~\citep{fu2023gptscore,liu2023gpteval}, building on their impressive reasoning capabilities. Despite efforts to revise factual errors in the text, a standardized evaluation metric has been lacking. Consequently, alternative methods have been proposed to assess revised text quality, aiming to replace costly human evaluation. Approaches like FActScore~\citep{min2023factscore} and SAFE~\citep{wei2024longform} focused on evaluating LLM-generated text factual accuracy, but still relied on external sources which is potentially  inaccurate. In contrast, our work used human-annotated data from FActScore to directly measure correction accuracy and revision accuracy.

\vspace{-3mm}
\section{Conclusion}
\vspace{-1mm}
In this paper, we propose \ours{} which detects and revises the factual errors in LLM-generated texts. The key idea of \ours{} is revising after explaining the factual error, based on the evidences gathered by external sources. Experimental results on GPT-3.5 and GPT-4 for multiple datasets show that \ours{} has a higher accuracy in detection and revision of factual errors, at lower cost. The demonstrated efficacy of \ours{} underscores its potential to foster trust and credibility in LLM-generated content across various domains, paving the way for further advancements in the field of artificial intelligence and natural language processing.

\section*{Limitation}
\vspace{-1mm}
\paragraph{Limitations of Evaluation Metrics.} 
While our evaluation metrics for revision task, \textit{correction accuracy} and \textit{revision accuracy}, are designed to evaluate the performance of \ours{} and baseline models in revising responses with factual errors, they have certain limitations. 
The \textit{correction accuracy} measures the proportion of false fact units in the initial response ($R_{\text{initial}}$) that are correctly revised in the revised response ($R_{\text{revised}}$). However, this metric does not account for the quality of the corrections made. For instance, a response could be deemed `corrected' if the false information is merely deleted or negated without being replaced with accurate information. Such revisions, while technically reducing the number of false fact units, is not enough for being the revised response informative. 
The \textit{revision accuracy} evaluates the ratio of correct fact units in the revised response. However, due to the dataset provided in ~\citep{min2023factscore} comprising a higher proportion of true fact units (approximately twice as many as false ones), this metric might inadvertently favor models that simply retain the true facts while inadequately addressing the false ones.
These limitations suggest that while our metrics provide a quantitative measure of a model's revision capabilities, they might not fully capture the qualitative aspects of proper fact correction. Future research should consider developing more comprehensive metrics or methodologies that evaluate not only the factual accuracy but also the informativeness and the contextual completeness of model-generated revisions.

\vspace{-1mm}
\paragraph{Limitations of Our Method.}

While our method shows promising results, particularly with long-form datasets such as FactPrompt \citep{min2023factscore}, we could not achieve satisfactory results in the context of short-form text verification (\eg FEVER \citep{thorne2018fever}). It turns out that decomposing short text and retrieving relevant information is a challenging task for our method. In these cases, including evidences during the explanation and revision phases frequently resulted in information overload, which caused the model to lose focus on correcting factual inaccuracies. For instance, the revised output might become verbose or contain contradictory statements. These observations underscore the importance of evidence retrieval and interpretation, and thus suggest directions for future work on filtering the evidences.

\bibliography{iclr2024_conference}

\appendix
\section*{Appendix}

\section{Details}

\subsection{Details on the Baseline}
\label{app:baseline}

\begin{table}[h]
    \centering
    \small
    \begin{tabular}{@{}rccccccc@{}}
    \toprule
        &\multicolumn{1}{c}{\textbf{Evidence}}
        &\multicolumn{2}{c}{\textbf{Query Generation}}
        &\multicolumn{2}{c}{\textbf{Detection}} 
        &\multicolumn{2}{c}{\textbf{Revision}} \\

        \cmidrule(r){2-2}
        \cmidrule(r){3-4}
         \cmidrule(r){5-6}
         \cmidrule(r){7-8}


         \small{Methods} 
         & \multicolumn{1}{c}{\small{source}}   
         &\multicolumn{1}{c}{\small{level}}
         & \multicolumn{1}{c}{\small{$n$-shot}}
                
         &\multicolumn{1}{c}{\small{level}}
         &\multicolumn{1}{c}{\small{$n$-shot}} 
         &\multicolumn{1}{c}{\small{level}} 
         &\multicolumn{1}{c}{\small{$n$-shot}}  \\
         \midrule
        FacTool  & external&   claim   & 3-shot  & claim &  3-shot  & - & -\\
        RARR & external & response& 6-shot    & query & 6-shot & response  & 6-shot \\

        CoVE \textit{(F)}  & internal& response & 3-shot    & - & - &  response & 3-shot \\
        CoVE \textit{(F+R)} & internal & response   & 3-shot & query  & 3-shot  & response & 3-shot\\
        \midrule
        \textbf{\ours{}}  & external& response  & 0-shot  & response  & 0-shot & response & 0-shot\\
        \bottomrule
    \end{tabular}
    \caption{Comparison of baseline methods with our method \ours{} for evidence source, query generation, detection, and revision. The table shows the source of evidence (internal knowledge of language models or external search tools), the input/output level (claim, query, raw response, or response), and whether each step employs few-shot or zero-shot approaches across different methods.}
    
    \label{tab:baseline-comparison}
\end{table}

As shown in Table \ref{tab:baseline-comparison}, FacTool and RARR both rely on external search tools: Serper API \footnote{\url{https://serper.dev}} and Bing Search API \footnote{\url{https://www.microsoft.com/en-us/bing/apis}}, respectively, for evidence retrieval. In contrast, CoVE revises responses internally within the LLM without external searches. Official code implementations exist for FacTool and RARR. We implemented CoVE following the paper's description and the prompt provided in the appendix, due to the lack of official code. For standardization, all experiments were conducted using both GPT-3.5 ('gpt-3.5-turbo') and GPT-4 ('gpt-4') with a temperature setting of 0.

The details of the baselines and our evaluation methods for these baselines are as follows:

\begin{itemize}
    \item CoVE (Factored)~\citep{dhuliawala2023chainofverification} first generates the verification questions and answers one by one, uses these verification question and answer pairs as evidence to revise the response. Since this method only focuses on revising the response, it was evaluated solely on the revision task.
    
    \item CoVE (Factor + Revise)~\citep{dhuliawala2023chainofverification}, which is similar to CoVE(Factored) but adds a step before delivering the final revised response. This step involves iterating through each verification question and its answer, and cross-checking the consistency between the verification question(query) and the answer pair and the initial response, categorizing them into three labels: {\textit{'consistent'}}, {\textit{'partially consistent'}}, and {\textit{'inconsistent'}}. If the consistency for even one of the verification questions and answers is labeled as \textit{'partially consistent'} or \textit{'inconsistent'}, the initial response is considered false; otherwise, it is considered true for the evaluation of the detection task. 
    
    \item RARR~\citep{gao2023rarr} utilizes two models, the \textit{agreement model} and the \textit{edit model}, to check the agreement between the evidence and the initial response, then revises it only if disagreement is detected. Similar to CoVE(Factor + Revise), the output from the \textit{agreement model} is used for the evaluation of the detection task. If the output for even one piece of evidence is \textit{'disagree'}, the initial response is considered false; otherwise, it is considered true.
    
    \item FacTool~\citep{chern2023factool}, which categorizes the initial response as either true or false. Therefore, the response-level factuality label from the model is used directly. Since this method focuses only on detecting factual errors in the response, it was evaluated solely on the detection task.
    
\end{itemize}

\subsection{Details on the Dataset}
\label{app:dataset}

We pre-processed each dataset to meet the specific requirements of our experiments, which used response-level binary labels \{ true, false \}.  Table ~\ref{tab:dataset} provides descriptions of each dataset after pre-processing.

\begin{table}[h]
    \centering
    \small
    \setlength\tabcolsep{4pt}
    \label{experimental-result}
    \begin{tabular}{@{}rrccccc@{}}
        \toprule
        Dataset & level & \textbf{\# True} & \textbf{\# False}  & \textbf{\#  Total} \\
         \midrule 
         Annotated Dataset from FActScore  & fact-unit & 3194 & 1692  & 4886\\
         
         Annotated Dataset from FActScore & response & - & 157 &  157 \\ 
       
         FactPrompt  & response & 23 & 27 & 50  \\ 
         WiCE (test) & content & 111  & 247& 358 \\

        \bottomrule
    \end{tabular}
    \caption{Dataset Descriptions}
    \label{tab:dataset}
\end{table}

\begin{itemize}
\item  Annotated dataset from FActScore~\citep{min2023factscore}, contains three labels \{ \textit{S}, \textit{NS}, \textit{IR} \} for each fact unit, which is a short sentence conveying one piece of information. We excluded the \textit{'IR (Irrelevant)'} responses to simplify the dataset to two primary labels. Consequently, we categorized \textit{'S (Supported)'} as true, and \textit{'NS (Not Supported)'} as false. Additionally, if even a single fact unit within a response was false, the corresponding response was considered false. Otherwise, it was considered true.

\item WiCE~\citep{kamoi2023wice} dataset, it contains three categories \{ \textit{S}, \textit{PS}, \textit{NS} \}. For our purposes, \textit{'S (Supported)'} is considered true, while both \textit{'PS (Partially Supported)'} and \textit{'NS (Not Supported)'} are treated as False in our evaluation metrics. 

\item FactPrompt~\citep{chern2023factool} dataset provides the response-level labels in two categories \{ True, False \}
, which were used directly in our experiment without any further modification. 
\end{itemize}

\section{Revision Score \& Correction Score}\label{sec:rev-corr-score-details} 

Here we explain how $n_{ft}$ and $n_{tt}$ given in Sec.~\ref{sec:exp-setup} are measured by using NLI model introduced in~\citep{liu2023evaluating}.
Let $p$ be a fact unit, and let $R_{\text{revised}}$ be the revised response. Then, we check the output of NLI model, denoted by 
\begin{align*}
    q = \text{NLI}(p, R_{\text{revised}}) \in \{ \text{entails}, \text{neutral}, \text{contradicts} \}.
\end{align*}
Then, $n_{ft}$ is the number of fact units $p$ that satisfies two conditions: (1) the fact unit is incorrectly stated in the initial response $R_{\text{initial}}$ and (2) the output of NLI model satisfies $q \in \{\text{neutral}, \text{contradicts} \}$.
Similarly, $n_{tt}$ is the number of fact units $p$ that satisfies two conditions: (1) the fact unit is correctly stated in the initial response $R_{\text{initial}}$ and (2) the output of NLI model satisfies $q \in \{\text{entails} \}$.

\vspace{50mm}
\section{Output Examples}

\subsection{Response Revision Examples}

As shown in Table ~\ref{tab:revision-example1} and ~\ref{tab:revision-example2}, \ours{} successfully revised the factual error by replacing it with correct information, whereas other methods either failed to catch the error or simply negated it. 

\begin{table}[h]
\small
\centering
    \begin{tabular}{@{}rp{11cm}@{}}
        \toprule
        & \textbf{Initial Response} \\
        \midrule
        \textbf{GPT-3.5} & Lanny Flaherty is an American actor born on \textbf{December 18, 1949, in Pensacola, Florida}... Some of his notable film credits include \textbf{"King of New York"}, ... \\
        \midrule
        & \textbf{Revised Response} \\
        \midrule
        
        \textbf{RARR} & Lanny Flaherty is an American actor born on \textcolor{blue}{July 27, 1942, in Pontotoc, Mississippi} \\
        \vspace{1.5mm}
        & ... Some of his notable film credits include \textcolor{red}{"King of New York," ...} \\
        
        \textbf{CoVE} \textit{(F)} & Lanny Flaherty is an American actor. He was \textcolor{red}{not born on December 18, 1949}.\\ 
        \vspace{1.5mm}
         & ... He has appeared in films such as \textcolor{red}{"King of New York,"} ... \\
        \textbf{CoVE} \textit{(F+R)} & Lanny Flaherty is an American actor born on \textcolor{red}{December 18, 1949, in Pensacola, Florida}. \\
         &... He has appeared in films such as \textcolor{red}{"King of New York,"} ...\\
        
        \midrule
        \textbf{\ours{}} \textit{(one-step)} & Lanny Flaherty is an American actor born on \textcolor{blue}{July 27, 1942, in Pontotoc, Mississippi}.\\
        \vspace{1.5mm}
         &  ... Some of his notable film credits include \textcolor{blue}{"Signs" (2002)}, ... \\
        \textbf{\ours{}} \textit{(two-step)} & Lanny Flaherty is an American actor born on \textcolor{blue}{July 27, 1942, in Pontotoc, Mississippi}. \\ 
         & ... Some of his notable film credits include \textcolor{blue}{"The Ballad of Little Jo" (1993)}, ... \\
        \bottomrule
    \end{tabular}
    \caption{Example Output of Revised Response Methods powered by GPT-3.5. The \textbf{Bold} Text contains factual errors in the initial response. \textcolor{blue}{Blue} text contains correctly revised information, while \textcolor{red}{Red} text contains incorrectly revised information or not revised. }
    \label{tab:revision-example1}
    
\end{table}

\begin{table}[h]
\small
\centering
    \begin{tabular}{@{}rp{11cm}@{}}
        \toprule
        & \textbf{Initial Response} \\
        \midrule
        \textbf{GPT-3.5} & The most valuable NFT ever sold is a digital artwork called \textbf{'Everydays: The First 5000 Days'} by the artist Beeple, which sold for \textbf{\$69 million} at a Christie's auction in March 2021       \\
        \midrule
        & \textbf{Revised Response} \\
        \midrule
        
        \textbf{RARR} & The most valuable NFT ever sold is a digital artwork called \textcolor{red}{'Everydays: The First 5000 Days'} by the artist Beeple, which sold for \textcolor{red}{\$69 million} at a Christie's auction in March 2021.\\
        \vspace{1.5mm}

        \textbf{CoVE} \textit{(F)} & The most valuable Non-fungible Token (NFT) is \textcolor{red}{'Everydays: The First 5000 Days'} by Beeple, which was sold for \textcolor{red}{\$69 million} at a Christie's auction in March 2021.\\ 
        \vspace{1.5mm}
        \textbf{CoVE} \textit{(F+R)} & The most valuable Non-fungible Token (NFT) as of 2021 is \textcolor{red}{CryptoPunk \#7804}, which was sold for 4200 Ether, equivalent to approximately \textcolor{red}{\$7.6 million} at the time of sale. However, the value of NFTs can fluctuate and new sales may have occurred.\\
        \midrule
        \vspace{1.5mm}
        \textbf{\ours{}} \textit{(one-step)} & The most valuable NFT ever sold is actually a digital artwork called \textcolor{blue}{'The Merge' by the artist Pak}, which was sold for \textcolor{blue}{\$91.8 million} on the NFT platform Nifty Gateway in December 2021.\\
        
        \textbf{\ours{}} \textit{(two-step)} & The most valuable NFT ever sold is \textcolor{blue}{'The Merge' by the artist Pak}. However, 'Everydays: The First 5000 Days' by the artist Beeple is one of the most valuable, having sold for \$69.3 million at a Christie's auction in March 2021.\\
        \bottomrule
    \end{tabular}
    \caption{Example Output of Revised Response Methods powered by GPT-4. The \textbf{Bold} Text contains factual errors in the initial response. \textcolor{blue}{Blue} text contains correctly revised information, while \textcolor{red}{Red} text contains incorrectly revised information or not revised.}
    \label{tab:revision-example2}
    
\end{table}

\newpage

\subsection{Retrieved Evidence Examples}
Table~\ref{tab:sub-answer-example} shows the evidences retrieved in Step 1 of \ours{}, either using the external search or the internal retrieval method. One can confirm that external search API enriches the method by leveraging up-to-date and accurate information. Compared to the  approaches solely reliant on the internal knowledge of LLMs, which may be outdated or inaccurate, integrating externally sourced data can help LLMs revise the initial response with rich and accurate information.

\begin{table}[hbt!]
\small
\setlength\tabcolsep{3pt}
\centering
    \begin{tabular}{@{}rp{10.5cm}@{}}
    \toprule
        \textbf{Sub-Question} \small{(Query)} &  Does Thomas Houghton currently play as a hooker for Hull FC in the Super League?\\
        \midrule
        \textbf{Sub-Answer} \small{(Result)} \\
        \midrule
         \vspace{1.5mm}
        \textbf{Internal} $\cdot$ GPT-3.5 & \underline{I'm sorry, but I am unable to provide real-time information} on specific players' current teams. I recommend checking the latest news or the official website of the rugby league team to find out which team Thomas Houghton currently plays for. \\

          \vspace{1.5mm}
          \textbf{Internal} $\cdot$ GPT4 & \underline{I'm sorry, I cannot provide real-time or current information.} It is recommended to check the official Hull FC website or the Super League website for the most up-to-date information on player rosters and positions.\\
            
          \textbf{External} $\cdot$ Google Search &  Danny Houghton (born 25 September 1988) is \underline{an English rugby league footballer} who plays as \underline{a hooker for Hull F.C. in the Super League}. He has played his entire career for his hometown club Hull FC. He is an England Knights international ...  \textit{(Source: Wikipedia)} \\
           & A member of the Super League Dream Team in 2016, Houghton finished as Super League’s top tackler in \underline{2011, 2013, 2015, 2016, 2017, 2019 and 2022} whilst featuring in the top three in 2012, 2014 and 2018... In the 2022 season, Houghton surpassed Brian Hancock to become Hull FC’s leading living appearance maker ... \textit{(Source: FC Official Website)} \\
          
        \bottomrule
    \end{tabular}
    \caption{Example of Sub-answer which is the result of a search query, comparing 1) instructing LLMs to respond to the query using their internal knowledge, which is denoted as \textbf{Internal}, and 2) using organic results, which means top search results from the Search API, denoted as \textbf{External}. GPT-3.5 and GPT-4 cannot provide the information correctly; however, the external search gets the top 2 organic results: one is from Wikipedia, and the other is from an FC website.}
    \label{tab:sub-answer-example}
    
\end{table}

\newpage

\section {Comparing Query Generation Steps}
\label{app:query}

In this section, we compare the query generation step for baselines (FacTool~\citep{chern2023factool}, CoVe~\citep{dhuliawala2023chainofverification} and RARR~\citep{gao2023rarr}) and \ours{}. FacTool uses a two-step process for query generation, beginning with the claim extraction and then generating individual queries for each extracted claim. CoVE uses a plan verification step that directly generates verification questions.  The RARR method employs a comprehensive question generation (CQGen) process,  which first samples three questions from the CQGen model and then combines these results to form a comprehensive set of queries. Lastly, \ours{}, utilizes a question-decomposition based approach for the sub-question generation, breaking down the text into sub-questions.

\begin{table}[h]
    \centering
    \begin{tabular}{@{}rcccccccc@{}}
    \toprule
        &\multicolumn{2}{c}{\textbf{Time} \small{(sec, ↓)}}
        &\multicolumn{2}{c}{\textbf{Token} \small{(K, ↓)}} 
        &\multicolumn{2}{c}{\textbf{\# Queries}} \\
        \cmidrule(r){2-3}
        \cmidrule(r){4-5}
        \cmidrule(r){6-7}

         \small{Methods} &\multicolumn{1}{c}{\scriptsize{GPT-3.5}} & \multicolumn{1}{c}{\scriptsize{GPT-4}} 
         &\multicolumn{1}{c}{\scriptsize{GPT-3.5}} & \multicolumn{1}{c}{\scriptsize{GPT-4}} 
         &\multicolumn{1}{c}{\scriptsize{GPT-3.5}} & \multicolumn{1}{c}{\scriptsize{GPT-4}}  \\
        \midrule 
        FacTool &  2.5 & 11.0 & 1.9 & 2.0 & 8.3 & 8.9 \\
        CoVE &  \textbf{1.0} & \textbf{2.6} & \underline{0.6} & \underline{0.6} & 2.6 & 2.4 \\
        RARR  & 3.9 & 11.5 & 1.8 & 1.5 & 5.5 & 5.6\\
        \midrule
        \textbf{\ours{}} & \underline{1.6}  & \underline{8.9} & \textbf{0.2}  & \textbf{0.2} & 4.9 & 5.1  \\
        \bottomrule
    \end{tabular}
    \caption{Experimental results for the query generation step using the FactPrompt Dataset, comparing the average number of tokens, time, and the average number of generated queries for each method. Full step results are provided in Table ~\ref{tab:experiment2}.}
    
    \label{tab:ablation2}
\end{table}

As shown in Table~\ref{tab:ablation2}, \ours{} requires the fewest tokens compared to the other baselines. In terms of the number of queries and time, CoVE requires and generates the least, followed by \ours{}. As illustrated in Table~\ref{tab:query-example}, CoVE indeed generates fewer queries, which leads to poorer performance, as shown in Table~\ref{tab:experiment2}. In the case of FacTool and RARR, as seen in Table~\ref{tab:experiment2} and Table~\ref{tab:ablation2}, \ours{} uses fewer tokens and time while achieving better performance than both. Since FacTool and RARR do not generate queries in a single step, duplicated queries, as shown in Table~\ref{tab:query-example}, are often observed, leading to inefficiency.

\begin{table}[t]
\centering
\small{
    \begin{tabular}{@{}rp{12cm}@{}}
        \toprule
        & \textbf{Prompt} \\
        \midrule
        User & Who is the CEO of Twitter?\\
        \midrule
        & \textbf{Initial Response} \\
         \midrule
        GPT-3.5 & The CEO of Twitter at the time of writing this answer is Jack Dorsey. \\
        & He co-founded Twitter in 2006 and served as the CEO until 2008, and returned as CEO in 2015. \\
        & Dorsey is also the CEO and co-founder of Square, a financial services and mobile payment company. \\
        & He has been recognized as one of Time magazine's 100 most influential people in the world and has also been awarded the Innovator of the Year Award by Wall Street Journal. \\ 
        \midrule
         & \textbf{Generated Queries} \\
        \midrule
        FacTool & 1.  Who is the current CEO of Twitter? \\ 
        & \uline{2. Jack Dorsey} \\
        & 3.  Who co-founded Twitter in 2006?\\
        & \uline{4. Jack Dorsey}\\
        & 5.  When did Jack Dorsey serve as the CEO of Twitter?\\
        & \uline{6. Jack Dorsey CEO Twitter}\\
        & 7.  When did Jack Dorsey return as CEO of Twitter?\\
        & 8.  \uline{Jack Dorsey CEO Twitter 2015}\\
        & 9.  Who is the CEO of Square?\\
        & 10. Who is the co-founder of Square?\\
        & \uline{11. What is Square?}\\
        & \uline{12. Square financial services}\\
        & \uline{13. Has Jack Dorsey been recognized as one of Time magazine's 100 most influential people in the world?}\\
        & \uline{14. Time magazine 100 most influential people Jack Dorsey}\\
        & 15. Has Jack Dorsey been awarded the Innovator of the Year Award by Wall Street Journal?\\
        & 16. Jack Dorsey Innovator of the Year Award Wall Street Journal\\

        \midrule
        RARR & 1.  Who is the CEO of Twitter? \\ 
        & 2.  When was Twitter founded? \\
        & 3.  When did Jack Dorsey co-found Twitter? \\
        & \uline{4. What other company is Jack Dorsey the CEO of?} \\
        & \uline{5. What other company is Jack Dorsey the CEO and co-founder of?} \\ 
        & 6.  What other company does Jack Dorsey co-founded? \\ 
        & \uline{7. Has Jack Dorsey been recognized by Time magazine?} \\ 
        & \uline{8. Has Jack Dorsey been recognized by Time magazine and Wall Street Journal?} \\
        & 9.  Has Jack Dorsey been awarded the Innovator of the Year Award by Wall Street Journal? \\

        \midrule
        CoVE & 1.  When did Jack Dorsey co-found Twitter? \\
        & 2.  When did Jack Dorsey first become CEO of Twitter?\\
        
        \midrule
        \textbf{\ours{}} & 1.  Who is the CEO of Twitter?\\
        & 2.  When did Jack Dorsey co-found Twitter? \\
        & 3.  How long did Jack Dorsey serve as the CEO of Twitter in his first term? \\
        & 4.  When did Jack Dorsey return as CEO of Twitter?\\
        & 5.  What other company is Jack Dorsey the CEO and co-founder of? \\ 
        & 6.  What type of company is Square? \\ 
        & 7.  Was Jack Dorsey been recognized by Time magazine? If so, for what? \\ 
        & 8. What award has Jack Dorsey received from the Wall Street Journal? \\ 
        \bottomrule
    \end{tabular}
    \caption{Example of query generation output powered by GPT-3.5. Here, \uline{underlined} text contains duplicated queries. }
    \label{tab:query-example}
}
\end{table}

\clearpage

\section{Prompts}

\small
\label{app:prompt}
\begin{table}[h]
\centering
\small
\begin{tabular}{@{}p{14cm}@{}}
\toprule
\\
Your task is to decompose the text into simple sub-questions for checking factual accuracy of the text. Make sure to clear up any references.\\ 
\\
Topic: \texttt{[prompt]}\\
Text: \texttt{[response]} \\ 
\\
Sub-Questions: \\
\\
\bottomrule
\end{tabular}
\vspace{5pt}
\caption{Prompt template used for Sub-Question Generation.}
\label{tab:query-generation}
\end{table}

\begin{table}[h]
\centering
\small
\begin{tabular}{@{}p{14cm}@{}}
    \toprule
    \\
  You will receive an initial response along with a prompt. Your goal is to refine and enhance this response, ensuring its factual accuracy. Check for any factually inaccurate information in the initial response.\\
  \\
  
  Use the provided sub-questions and corresponding answers as key 
  resources in this process.
  Sub-questions and Answers :  \\ 
  \texttt{[ sub-question 1  \& sub-answer 1 ]} \\ 
  \texttt{[ sub-question 2  \& sub-answer 2 ]} \\ 
  \texttt{[ sub-question 3  \& sub-answer 3 ]} \\ 
  \texttt{ ... } \\
   \\

  Prompt: \texttt{[prompt] }\\ 
  Initial Response: \texttt{[response] }\\ 
  \\

  Please explain the factual errors in the initial response, and revise it accordingly. \\ 
  If there are no factual errors, respond with  "None". \\ 
  If there are factual errors, explain each factual error.\\ 
  \\
  Factual Errors: \\
  Revised Response: \\
  \\
\bottomrule
\end{tabular}
\caption{Prompt template used for \ours{} \textit{(one-step)}: Step 2 \& 3. Factual Error Explanation \& Revision.}
\label{tab:one-step}
\end{table}

\newpage

\begin{table}[h]
\centering
\small
\begin{tabular}{@{}p{14cm}@{}}
    \toprule
    \\
  You will receive an initial response along with a prompt. 
  Your goal is to refine and enhance this response, ensuring its 
  factual accuracy. Check for any factually inaccurate information 
  in the initial response. \\
  \\
  
  Use the provided sub-questions and corresponding answers as key 
  resources in this process.
  Sub-questions and Answers :  \\ 
  \texttt{[ sub-question 1  \& sub-answer 1 ]} \\ 
  \texttt{[ sub-question 2  \& sub-answer 2 ]} \\ 
  \texttt{[ sub-question 3  \& sub-answer 3 ]} \\ 
  \texttt{ ... } \\
   \\

  Prompt: \texttt{[prompt] }\\ 
  Initial Response: \texttt{[response] }\\ 
  \\

  Please explain the factual errors in the initial response.\\ 
  If there are no factual errors, respond with  "None". \\ 
  If there are factual errors, explain each factual error.\\ 
  \\
  Factual Errors: \\
  \\
\bottomrule
\end{tabular}
\caption{Prompt template used for \ours{} \textit{(two-step)}: Step 2. Factual Error Explanation.}
\label{tab:explanation}
\end{table}

\begin{table}[h]
\small
\centering
\begin{tabular}{@{}p{14cm}@{}}
\toprule
\\
  You will receive an initial response along with a prompt. Your goal is to refine and enhance this response, ensuring its factual accuracy. Check for any factually inaccurate information in the initial response. You will receive a list of factual errors in the initial response 
  from the previous step. Use this explanation of each factual error 
  as a key resource in this process :\\
  \\
  Factual Errors:\\
  \texttt{[ explanation-of-factual-error 1 ]}\\
  \texttt{[ explanation-of-factual-error 2 ]}\\
  \texttt{[ explanation-of-factual-error 3 ]}\\
  \texttt{...}\\
  \\
  Prompt: \texttt{[prompt]}\\
  Initial Response: \texttt{[response] }\\
  Revised Response: \\
  \\
  
\bottomrule
\end{tabular}
\caption{Prompt template used for \ours{} \textit{(two-step)}\: Step 3. Revision.}
\label{tab:revision}
\end{table}

\newpage

\end{document}